\documentclass{article}
\usepackage{graphicx} 
\usepackage{authblk} 
\usepackage{breakurl} 
\usepackage{geometry}
\usepackage{hyperref}
\usepackage[utf8]{inputenc} 
\usepackage[T1]{fontenc}    
\usepackage{hyperref}       
\usepackage{url}            
\usepackage{booktabs}       
\usepackage{amsfonts}       
\usepackage{nicefrac}       
\usepackage{microtype}      
\usepackage{xcolor}         
\usepackage{graphicx} 
\usepackage{amsmath}
\usepackage{amssymb}
\usepackage{graphicx}
\usepackage{natbib}
\usepackage{caption}

\DeclareCaptionFormat{myformat}{\fontsize{8}{10}\selectfont\textbf{#1}#2#3}
\begin{document}
\title{Towards Signal Processing In Large Language Models}

\author{Prateek Verma}
\author{Mert Pilanci}
\affil{Department of Electrical Engineering \\
     Stanford University\\
     Stanford, CA, 94305}

\date{May 2024}

\maketitle

\begin{abstract}
This paper introduces the idea of applying signal processing inside a Large Language Model (LLM). With the recent explosion of generative AI, our work can help bridge two fields together, namely the field of signal processing and large language models. We draw parallels between classical Fourier-Transforms and Fourier Transform-like learnable time-frequency representations for every intermediate activation signal of an LLM. Once we decompose every activation signal across tokens into a time-frequency representation, we learn how to filter and reconstruct them, with all components learned from scratch, to predict the next token given the previous context. We show that for GPT-like architectures, our work achieves faster convergence and significantly increases performance by adding a minuscule number of extra parameters when trained for the same epochs. We hope this work paves the way for algorithms exploring signal processing inside the signals found in neural architectures like LLMs and beyond. 
\end{abstract}

\section{Introduction and Related Work}
We are currently in the middle of an AI super-renaissance. A significant driver of this has been Transformer architecture \cite{vaswani2017attention}. Initially designed for machine translation, it has also become a core backbone in computer vision \cite{dosovitskiy2020image}, audio \cite{verma2021audio}, robotics \cite{brohan2022rt}, statistics \cite{nie2022time}, computational biology \cite{madani2020progen}, music generation \cite{huang2018music} to name a few. One of the main reasons for this success is that Transformer architecture was designed for solving seq-to-seq tasks, and most of the problems around us can be posed as one/many sequence to one/many sequence mapping tasks, even a problem like object detection \cite{chen2021pix2seq}. 
\begin{figure*}[t]
  \centering
  \hspace*{8.8pt}
  \includegraphics[width=\linewidth,height=9cm]{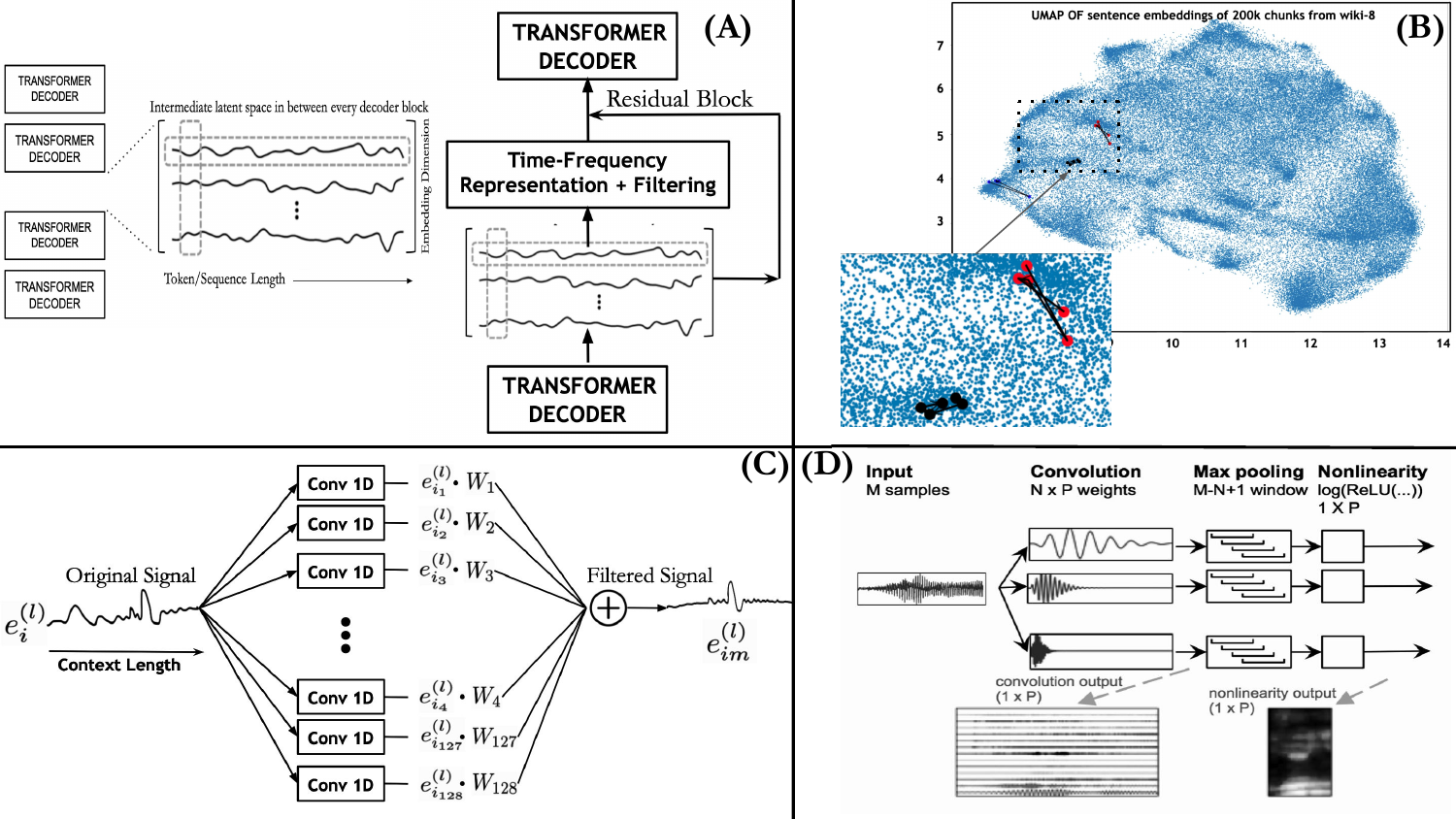}
  \caption{(A) Incorporating the idea of doing signal processing inside a Large Language Model. We find signals between every decoder layer across token dimensions as 1-D signals. Each of these signals is then decomposed into a time-frequency representation and filtered and added back akin to a residual block (B) Path interpretation, where our method can be interpreted as filtering the latent space of the path of how intermediate embeddings of text tokens traverse in high-dimensional space. We can see embeddings with black marker follow circular path and constrained to certain regions. We design filters to learn these paths and constraint them or shape the trajectory.  (C) We compute a time-frequency representation for each signal via 1-D causal filters. Each of the decomposed signals, with a relu non-linearity is then \textit{amplified/supressed} by learned weights and added back to original signal to get the filtered signal (D) Figure from \cite{sainath2015learning}. We draw inspiration from learning a time-frequency representation used in speech recognition that learns a mel-spectrogram-like representation from scratch using convolutional filters rather than a Fourier-based representation. }
  \label{fig:decoder_causal}
\end{figure*} 
There has been previous work in a similar theme of integrating signal processing algorithms with deep neural networks. Differentiable Digital Signal Processing \cite{engel2020ddsp} mimicked a traditional signal processing pipeline for various audio transformation and synthesis tasks. The work showed how, with a guided choice of learnable parameters, we can achieve interpretable, powerful results with limited non-aligned data, which otherwise would not have been possible with having an abundance of data. The integration of signal processing ideas such as Fourier Transform has also been explored recently in FNet \cite{lee2021fnet}. This work in non-causal setups such as BERT \cite{devlin2018bert} replaced attention blocks with that of taking 2-D Fourier Transforms. This is done by combining two 1-D Fourier Transforms and replacing the attention block with it. However, taking FFT 12 times (for a 12-layer model) makes the approach uninterpretable. It cannot be explained via any of the classic signal processing literature or textbook \cite{oppenheim1975digital} to our knowledge. It is also only defined for non-causal setups and, thus, cannot be adapted to causal architectures at the core of LLM pre-training. Thus, the gains cannot be extrapolated to generative model and language modeling tasks such as GPT \cite{brown2020language}. The work that closely resembles our work is Language through Prism \cite{tamkin2020language}. This paper, again designed for non-causal setups, proposes that embeddings across tokens can be fine-tuned at different rates for downstream tasks like topic modeling and part-of-speech tagging. They achieve this by decomposing the intermediate embeddings across token coordinates through a discrete cosine transform \cite{ahmed1974discrete}. They built hand-tuned heuristics for natural language tasks, e.g., for document-level classification for a collection of words, we only need to learn linear probe over final layer embeddings moving slowly (filtered by low pass filter over token dimension), whereas for tasks such as part of speech tagging, where word-level prediction is needed,  spectral filters designed were high pass filters. The paper also presented a prism layer for training BERT-like models, which used spectral filters to constrain the neural activation of neurons at different scales. A significant drawback of this work is that it uses a pre-defined time-frequency representation, namely Discrete Cosine Transform which may be sub-optimal for the space of activations of neural architectures \cite{sainath2015convolutional,verma2023content}. Besides this, it was only proposed for non-causal architecture such as BERT, as Discrete Cosine Transform was computed across the context length of all the tokens. This cannot be thus extended to the Large Language Model training setup, where we predict the next token causally for every token. Finally, a subtle difference in their work is that the choice of weighing function is fixed or computed via hand-made heuristics. A high-pass or low-pass filter is achieved by turning off specific frequency bins and turning on other bins. What if the filter needed to have complex pass-band or stop-band characteristics? In our work, we devise an architecture that can learn an optimal time-frequency representation from scratch, which retains the causality assumption. We also learn a weighted sum on learned time-frequency estimation in which the weights are not boolean but continuous and learned according to the optimization criteria, allowing us to turn on or off the time-frequency representation learned. All of this is learned with a single objective function driving these systems: What is the best time-frequency representation followed by filtering to optimize for the next token prediction? 
\footnote{The Figure 1 (B) is computed as follows: We take sentence embedding for chunks of 128 characters to extract embeddings of 200k chunks from wiki-8 dataset. It is clustered using UMAP algorithm. We take 3 new paragraphs from wikipedia for testing and extract 5 chunks from them. They are shown as blue, red, and black respectively taken from first few sentences of online wikipedia article of "Fourier Transform", "Penguin" and "Solar Eclipse" respectively} 
Our paper will also explore an alternate line of work: learning time-frequency representations from ideas in speech and audio processing while retaining the causality assumption. Typically, Fourier Transform is used to decompose a time-domain signal into its constituent signals\cite{SASPWEB2011}. The first step for a typical audio processing pipeline is to compute a Fourier Transform-based STFT as the signal varies across time. Spectrograms are typically used to find an image-like 2-D representation that can enable learning-based algorithms like CNN, Transformers, and GPT to be used for problems such as classification \cite{hershey2017cnn,verma2020framework,gong2021psla}. However, they are suboptimal representations \cite{zeghidour2021leaf} e.g., i) All basis functions are fixed as sinusoidal. What if, for specific applications, we need a different choice of basis function \cite{sainath2015learning,verma2016frequency}? Secondly, they are spaced equidistant across the frequency range. iii) They are sub-optimal for the choice of problem, i.e., for different problems, we learn different time-frequency representations optimized for the problem at hand \cite{ravanelli2018speaker}. In the early 90s, different ways of mapping Fourier Transform to how humans hear sounds gave rise to the mel/bark scale, which closely matches the psycho-acoustics of human hearing \cite{stevens1937scale}. This representation is still quite successful and used ubiquitously for ASR/audio signal processing \cite{,hershey2017cnn,wang2017tacotron}. However, modern neural architectures have given rise to the representation that can be learned according to the task \cite{verma2021audio}, the input data distribution, e.g., a different representation for noisy vs. clean signal \cite{sainath2015learning} or the input content itself \cite{verma2023content}. In the next section, we will motivate this choice and figure out how to make primitive signal processing applications, such as filtering inside of LLM, during pre-training. We will also address whether this helps and further improve our proposed methods with even more complex methods, such as learning optimal time-frequency masks that vary across tokens. 
\newline
\newline
The contribution of the paper is as follows: 
\begin{itemize}
\item We show how to incorporate a core signal processing idea, filtering, into LLM pre-training. We draw parallels between learning time-frequency representation in a causal manner and filter-bank representation of a Fourier Transform. We find signals in the intermediate embeddings inside GPT architecture. We decompose these signals into constituent components by a learnable time-frequency representation, filter in this learned representation, and reconstruct the signal. These are optimized and driven just by the next token prediction. 
\item We further bring in complexity and mimic learning time-frequency mask that changes the characteristics of the learned filter as opposed to fixed filters across the token dimensions. We show that the method of filtering using time-frequency masks, initially developed for source separation in audio signals, can be incorporated into LLM pre-training and further improve the performance of pre-training LLM.

\item For non-causal setup, we take a simple audio classification task and show how our method can also be adapted for non-causal setups. We take signals in intermediate embedding dimensions along tokens, compute the Discrete Cosine Transform of the signals to decompose the signal into its constituent components, learn from scratch filters on the DCT coefficients to filter the signal, and reconstruct the signal back. The pipeline is learned from end-to-end, with the classification loss function driving the whole system. We show improvements in classification accuracy with a baseline architecture not having such as a pipeline. 
\end{itemize}
\section{Dataset} We utilize publicly available datasets to showcase the strength of our work. For natural language processing, we use text-8 dataset \cite{text8,mikolov2012subword}. We follow character-based tokenization similar to \cite{al2019character} to make the data preprocessing as simple as possible. This simple tokenization allows us to use a vocabulary of only 27 possible tokens that can be predicted in an LLM setup. It mitigates the extra tokens like digits by spelling them out. All other symbols, like punctuations, are replaced by a single space. All of the characters are also converted to lowercase characters, thus allowing to have 26 characters + space as possible vocabulary. We thus remove the effects of different tokenizers and focus purely on the effect of architecture. We used setup same as \cite{al2019character}: we split the data into 90M characters for train, 5M characters for dev, and 5M characters for test as given in \cite{al2019character}. All of the results reported are on the test set before and after doing signal processing based processing on intermediate GPT-like embeddings. For classification results that we report for audio signals, we use FSD-50K dataset \cite{fonseca2020fsd50k}, containing about 51k audio snippets at 16kHz, for 200 categories of sound, sharing the same ontology as AudioSet \cite{gemmeke2017audio}.

\section{Background} This section will define our problem and introduce the Fourier Transform and the background needed for our work. We will explain the filter-bank interpretation of the Fourier Transform and how a convolution operation can approximate it. Further, we show how to connect these concepts to make the operation causal and suitable for  LLM pre-training which predicts the next token given context. 

\subsection{Fourier Transform Preliminaries} Given any signal $x(n)$ with length $N$, Discrete Fourier Transform DFT transforms the sequence of its constituent into a sequence of $N$ complex numbers $X(\omega_k)$ which is defined as,

$$X(\omega_k) = \sum_{n=0}^{N-1} x(n) e^{-i\omega_kn}; X(k) = \sum_{n=0}^{N-1} x(n) \cos\left[\frac{\pi}{N} \left(n + \frac{1}{2}\right) k \right]$$

where $\omega_k$ is defined as $2\pi k/N$. We can see from this definition that the magnitude of each of the outputs in the sequence $X(\omega_k)$ corresponds to the strength of the presence of a particular band of frequencies. The definition is a way of decomposing a signal into a complex sum of sinusoidal basis functions. There are variants of this, including Discrete Cosine Transforms DCT \cite{ahmed1974discrete}, which can operate on real-valued signals and transform them into a real-valued sequence. We will show our approach can be adapted for classfication tasks, which is a non-causal operation, and can operation on DCT coefficients. Given a $x(n)$, we define DCT-II of the signal as, taking the signal of length $N$, and transforming it to a $N$ length sequence $X(k)$ which is defined in the equation above. 
\subsection{Learnable Time-Frequency Representation} There however exist several drawbacks in this classical defination of DFT/DCT as described in \cite{verma2016frequency,zeghidour2021leaf, ravanelli2018speaker}. This choice of basis functions may not be optimal for the type of problem or the dataset onto which the model is being trained. Another drawback is that the frequency spacing is linear, with equal allocation from lowest to highest frequency, with constant bandwidth. \cite{verma2016frequency} showed that if we learn a real-valued transform from scratch via multiplicative kernel, we see that the learned transform learns a non-linear, non-constant bandwidth kernel that is optimal for a given task. Similar approaches have been explored widely across audio signal processing and speech recognition literature to improve the performance of the task at hand. \cite{zeghidour2021leaf, ravanelli2018speaker,sainath2015learning}. With the advent of modern AI architectures, much work has been carried out on learning the optimal time-frequency representation that is not only adaptive according to the task \cite{ravanelli2018speaker, zeghidour2021leaf}, the dataset, whether it is noisy or clean \cite{sainath2015learning} or the content itself \cite{verma2023content}, yielding significant gains in the performance for downstream tasks. Motivated by this line of research, explore learning time-frequency representation that is optimal according to the data and the task for the next token prediction in a space spanned by the intermediate embeddings of an LLM for all intermediate layers.

\subsection{Short Time Fourier Transform As Filter-Bank}
Typically, a signal does not remain stationary but varies across time (or token, in our case). In these setups, instead of taking the Fourier Transform, it is common in signal processing and speech applications to take Short Time Fourier Transform (STFT). An STFT is taking a continuous signal, windowing it into patches, and computing the DFT at the patch level. It is defined as,

$$X_m\left(\omega_k\right)=\sum_{n=-\infty}^{\infty}[w(n-m) x(n)] e^{-j \omega_k n} = \sum_{n=-\infty}^{\infty} \underbrace{\left[x(n) e^{-j \omega_k n}\right]}_{x_k(n)} w(n-m) = \left[x_k * \operatorname{FLIP}(w)\right](m) $$

Where $w$ is a windowed signal that is non-zero at a finite interval, the definition is similar to taking the DFT of the modified signal \cite{allen1977unified}. An STFT also has a filter-bank interpretation. The terms can be rearranged as the output of a filter bank to give the same output. 
Now, "STFT is thus interpreted as a frequency-ordered collection of narrow-band time-domain signals, as depicted in Fig.9.2. In other words, the STFT can be seen as a uniform filter bank in which the input signal $ x(n)$ is converted to a set of $ N$ time-domain output signals  $ X_n(\omega_k)$,  $ k=0,1,\ldots, N-1$, one for each channel of the $ N$ -channel filter bank" \cite{SASPWEB2011}. This definition will help us explore what happens when we approximate computing the STFT of a signal with a learnable bank of convolutional filters (bank of filters). Let us take the magnitude of the signal above. We can see that a learned slice of an STFT (similar to a log-magnitude spectrogram) can be approximated by a learnable output of a convolutional filter as shown by \cite{sainath2015convolutional}. If we make these filters causal, we get time-frequency representation causal, which can be used in an LLM-like setup. 
\subsection{Filtering} Given a signal, filtering operation allows specific frequencies or components to pass through and certain components to be shunned. To give a context here, digital filter design is a field in itself, with textbooks written on the topic \cite{shenoi2005introduction}. However modern AI and deep learning, allow us to learn optimal filters for a given task, optimized for the task and the data at hand. Another way of looking at designing filters is analysis-synthesis approach \cite{princen1986analysis}-- to take a given signal, decompose it into its constituent components, and reconstruct the signal back from its constituent components by adjusting the weights/filtering each of the components. For designing a low-pass filter, we set all the frequency components above a certain frequency threshold, as off and allow only lower frequencies to pass through. It is highly relevant to our task that by finding signals inside large language models, we can first learn a time-frequency representation to decompose the signal into its constituent components, followed by learning how to optimally weigh each component (filter), and then modify the original signal to replace/add back to the original signal. All of these components, that is learning a time-frequency component and the filter on these signals are learned from scratch according to the optimization criterion at hand, which is the next token prediction given the context. This is the overall theme for our work and is explained in more detail in the next section.

\section{Methodology} We, in this section, explain the method that we proposed, with the the background explained in the previous section. We use a mini-GPT-like architecture as our baseline. For the sake of our paper, we do not compare our results with other state-of-the-art models due to the sheer size of the current LLM models \cite{brown2020language, fedus2021switch}. This paper was written in academia, with limited computational resources. Thus, we do the following setup: We train all models from scratch. All models are based on decoder-only Transformer architecture but scaled down in terms of the number of parameters by reducing the number of layers, embedding dimension, and number of heads. We do this by the following choice: The number of layers was chosen to be 8, with 128 as the model dimension and 512 as the feed-forward dimension and context length of 256. We use eight attention heads in every layer. The output of the last layer of the Transformer was then piped through a classification head (MLP layer with 2048 neurons followed by another dense layer equal to the vocabulary dimension) to map it to a number of vocabulary dimensions followed by a softmax function. We used cross-entropy loss as our loss metric and followed a typical LLM training recipe similar to GPT training. The learning rate was chosen to start from 3e-4. For text-8 the number of tokens that the model was trained on was about 1/2 billion. We then report improvements proposed to our method, relative to the baseline architecture in terms of the gain in validation loss and the speed of convergence in training steps. We believe that the ideas proposed by our paper can be applied to much larger, deeper models with little to no tweaks. In addition since we are doing processing of the intermediate embeddings, our work can be applied to non-transformer based LLM architectures such as MAMBA \cite{gu2023mamba}. 

\subsection{Finding Signals Inside Large Language Models} To do any signal processing, we first need to operate on signals. How can we find 1-D signals inside LLM? For this, we take a similar approach to that of \cite{tamkin2020language}, where we take signals across tokens. Our work uses decoder-only GPT-like architecture as our generative model for all experiments. For an architecture with $N+1$ Transformer only decoder blocks with context length $L$ and embedding dimension $E$, we define a signal across the entire context length on each coordinate of the embedding dimension. Thus, for a $N+1$ decoder block GPT architecture, we get $NE$ signals for length $L$. These are the signals on which we would do signal processing. A point to note is that since we are operating on a causality assumption, in order to prevent any leakage, we have to do any signal processing operation on these signals in a causal manner without being allowed to do any operation which looks at the future latent representations/signals. 

\subsection{Learnable Time-Frequency Representation And Filtering Over Signals} In this experiment, we apply a learnable time-frequency representation to intermediate embedding signals in all layers. A causal convolutional block mimics a time-frequency representation as motivated in the previous section. This is akin to learning an STFT, which can decompose the contents of the signal into its constituent components. (The only difference being the transform is learnable from scratch). Once we have decomposed the signal in this manner, we can learn to weigh which frequencies in this decomposed time-frequency representation are important or not important. 

In our experiments, as a common theme, we take each of the $NE$ signals as defined in the previous section, do some processing on it, and add it back to the original signal. As motivated in the section on connecting time-frequency representations and Fourier Transforms, following the idea of learning a front end for the audio signal, similar to \cite{sainath2015learning}. We opt for a simple learnable time-frequency representation via a 2-layer CNN architecture as follows (a set of conv filters followed by weighing): we use $M$ causal convolutional filters (chosen to be 144) to decompose the contents of the signal into its constituent components. In this experiment, we set the length of the convolutional filter to be 7. Once we take a particular the signal out of $NE$, for example after the $l^{th}$ layer in the $i^{th}$ dimension of the latent space $e_{i}^{(l)}$ of length $L$: After taking $M$ convolutional filters $h_{k}$, as seen in the equation with $\circledast$, being the convolution operation with $k \in [1,M]$, we get the output  signals as $e_{i_1}^{(l)}, e_{i_2}^{(l)}, e_{i_3}^{(l)}, ... e_{i_M}^{(l)}$. Each of these signals uses relu non-linearity after the output of the convolutional filter. A non-linear operation is important as traditional audio representations like MFCC uses a non-linear operation (log operation), and we also draw inspiration from learnable audio representation using relu's proposed in \cite{sainath2015learning}. As shown in \cite{sainath2015learning}, for our work the output of the first layer of convolutional filters can serve as a time-frequency representation that is learned at a particular point in the signal. We now learn scalars $w_{1},w_{2},w_{3} ....w_{M}$, which are used to amplify/weigh outputs of certain frequency components that is learned by the convolutional filters. The output we get after weighing is $f_{i}^{(l)}$, 
$$f_{i}^{(l)} = \sum_{k \in \{1,M\}} w_{k}  \max(0, h_{k} \circledast e_{i}^{(l)})  \hspace{1cm} g_{i}^{(l)} = \sum_{n \in \{1,4\}}  \sum_{k \in \{1,M/4\}} w_{k}  \max (0,  h^{(n)}_k \circledast e_{i}^{(l)})  $$

We then add this modified signal back to the original signal akin to residual networks \cite{he2020resnet}, $ e_{i}^{(l)} = e_{i}^{(l)} + f_{i}^{(l)} $. This operation can be loosely described as taking a signal, decomposing it into its frequency components, and learning (to design) a filter to turn weigh the output of these convolutional filters (frequency components). All of the components are learnable, i.e., the convolutional kernels and how to weigh these kernels, which  is optimized for next token prediction criteria using a cross-entropy loss. All of the modules are learned driven by the LLM next token prediction This is one of the most primitive and elementary signal processing operation that could have been done, i.e., filter design, which is at the heart of the field of signal processing. We will now describe the idea of learning multi-scale time-frequency representation, and designing advanced filters which can vary their characteristics across tokens as opposed to keeping the filter behaviour fixed as described here. 

\subsection{Multi-Scale Time-Frequency Representation And Filtering Over Signals} There have been papers in ASR literature \cite{zhu2016learning} on how a front end can be improved upon with the notion of multi-scale filters in the first layer. It was shown in this paper, that convolutional filters can be used to push past tradeoff of temporal and frequency resolution for learned spectral resolution. It yielded significant performance for speech recognition. Inspired by this idea, we incorporate this idea for the space of signals present in a large language models. We use convolutional filters with multiple resolutions, keeping the number of filters $M$ to be the same as 144. Here, we divide the filters into four blocks with convolutional filter lengths of 3, 7, 15, and 31, each utilizing $M/4$ filters. We again learn the weights of these convolutional filters and then the weighing scalar parameters $w_{1},w_{2},w_{3} ....w_{M}$ which are then used to weigh the output of the convolutional filters, followed by non-linearity similar to what was described in the previous section. The output we get after weighing is $g_{i}^{(l)}$. We add this modified signal to the original signal as described in the previous section, $e_{i}^{(l)} = e_{i}^{(l)} + g_{i}^{(l)} $

\begin{figure*}[t]
  \centering
  \hspace*{8.8pt}
  \includegraphics[width=\linewidth]{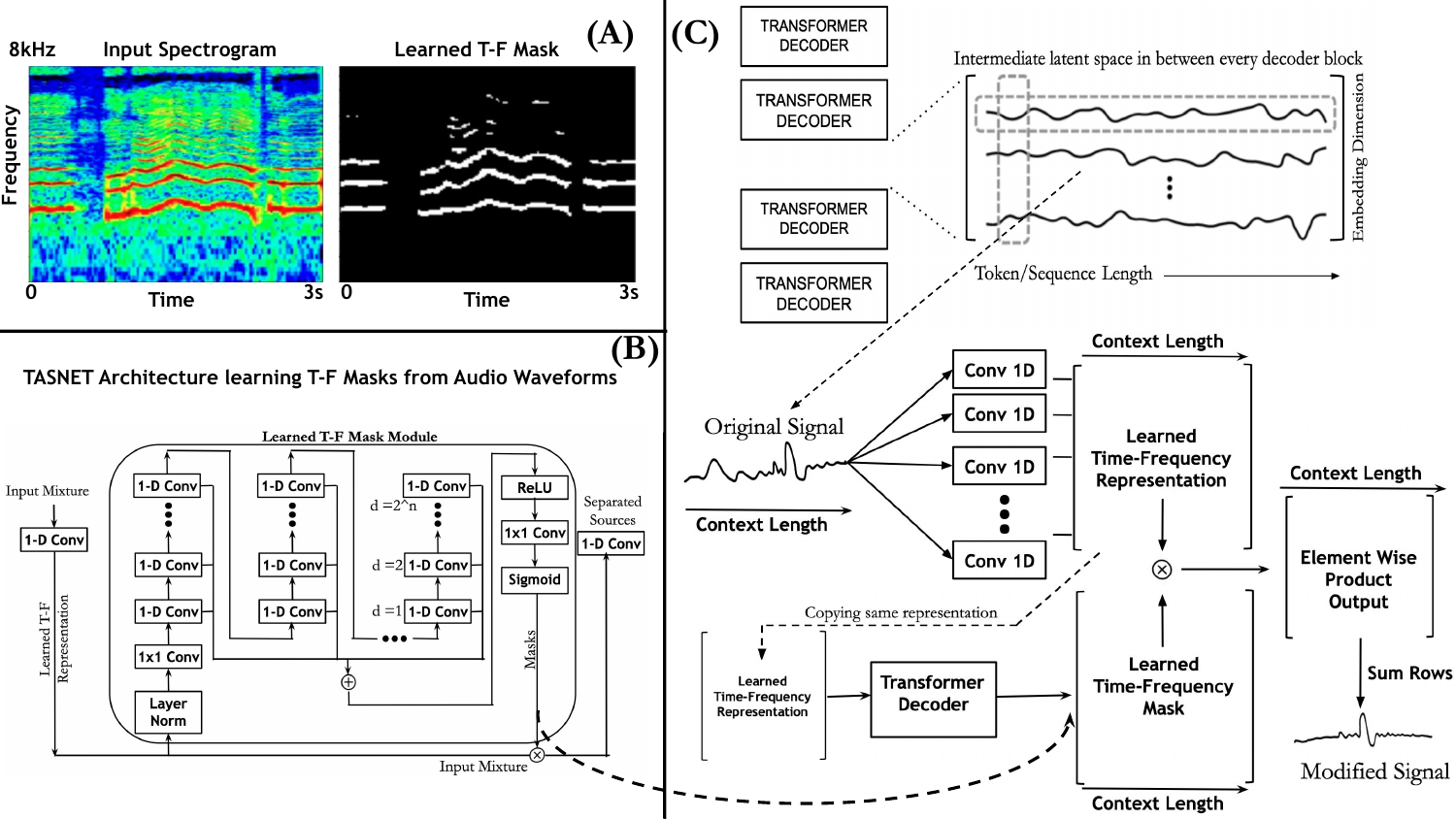}
  \caption{(C) We draw inspiration from a classic signal processing pipeline to bring advanced ideas in filtering to large language models. We take each of the original signals, decompose them into a time-frequency representation. We then learn a time-frequency mask on the learned representation, which allows us to learn time-varying filters as opposed to fixed filters across token dimension. (A) We have drawn inspiration from source separation pipeline, where spectrograms are taken, and time-frequency masked are learned to multiply the spectrogram representation to supress the signals were are not interested: Figure from \cite{reghunath2023predominant} (B) TASNET as shown in \cite{luo2019conv} proposed an architecture where a 1-D convolutional filter learns to decompose a signal and a series of convolutional  filters are used to learn a mask which is used to turn-on or off the decomposed signal and are multiplied together to retain the signal components  of interest. We can see parallels to our work in (C), where a Transformer decoder learns a t-f mask which allows the weights of the filter operating on the learned t-f representation varying across token/context dimension. }
  \label{fig:content-adaptive-filters}
\end{figure*} 

\subsection{Making Weights Adaptive Across Tokens}
We explore if we can make the weights adaptive across the tokens. We draw inspiration from a classic signal processing pipeline of source separation and learning time-frequency masks \cite{luo2019conv}. For primitive applications, filtering with fixed frequency bins turned-off/on works if the noise characteristics are stationary and limited throughout in some discrete frequency bands. However, the noise and signal of interest often change and disperse over multiple frequency bins as a function of time. A good example of this is source separation and the idea of time-frequency masking. The main idea is as follows: We first compute a spectrogram based time-frequency representation, and depending on the contents of the input log-magnitude spectogram, learn a mask which can turn-on/off specific time-frequency bins. This allows our filtering mask in the time-frequency representation to adapt according to the input signal characteristics. We apply the same analogy to that of the signals. Instead of having a constant filter characteristics across tokens as explained in previous section, we vary the filter-characteristics at every token. Mathematically, it becomes now 
$$ft_{i}^{(l)} = \sum_{k \in \{1,M\}} w^{t}_{k}  \max(0, h_{k} \circledast e_{i}^{(l)})$$

 where the weights itself is a function of the input signal namely $e_{i}^{(l)}$. We have to make sure that we retain the causality assumption,  hence in order to learn the weights, we pass on as an input the signal $e_{i}^{(l)}$ and learn as the output of the model, a signal that varies across token dimension $w^{T}_{k}$. This is carried out using a single layer Transformer decoder block, which is a light-weight one. It takes as inputs as the learned t-f representation projects it to a smaller dimension of 32 from $M= 144$ sized representation. The single layer Transformer decoder has 4 attention heads, and model dimension as 32. We then project the output of the Transformer back to $M$. The output of the Transformer are weights which are multiplied with that of the learned time-frequency representation. We impose causal mask on the decoder thereby retaining the causalilty assumption in the block that learns time-frequency masks. We compare the performance with/without making the filters time-varying as can be seen in Table 1 and see a jump in performance. 

\begin{figure*}[t]
  \centering
  \hspace*{8.8pt}  \includegraphics[width=0.8\linewidth, height=6cm]{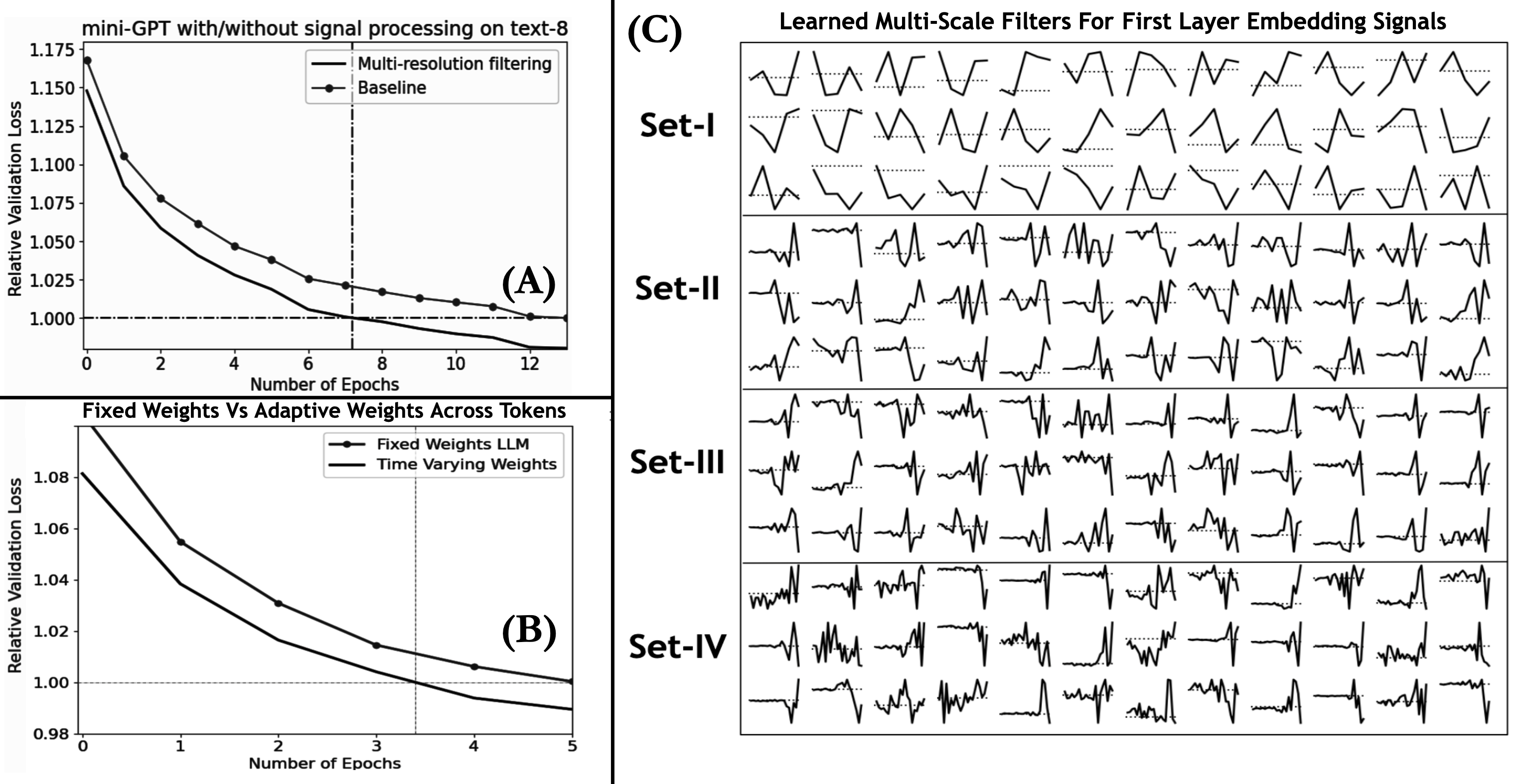}
  \caption{We compare performance of our proposed architecture that does signal processing on intermediate embeddings compared with that of baseline model for text-8. We can see that we achieve 40-45\% faster convergence. When we train it for the same duration of time, we see that we achieve close to 0.02 validation loss improvement.For text-8 trained LLM, we show how the filters look like for learned multi-scale time-frequency representation in the first layer. We see that the basis functions are not sinusoidal, emphasizing our hypothesis that we need to learn a time-frequency representation from scratch, which is optimal for the signals in the latent space between decoder layers. This plot is for the kernels learned after the first decoder layer.}
  \label{fig:relative-performance-plus-filters}
\end{figure*} 

\section{Results And Discussion}
\subsection{Performance In Non-Causal Setups} 

Before discussing performance in causal setups, we explored a non-causal setup: we manipulate intermediate signals across the token axis via simple non-learnable time-frequency decomposition of real signals, namely DCT-II. We take a simple toy classification task for audio signals for FSD-50K. We propose to operate in a non-causal setup here, where we take as an input the whole intermediate embedding at once, namely  $e_{i}^{(l)}$ where $l$ varies from 1 to the number of decoder layers, and $i$ varies for 1 to the embedding dimension $E$. Due to the non-causal setup, we can look into the future tokens to modify the current embedding coordinate. We, for every signal $e_{i}^{(l)}$, compute DCT-II for the signal length equal to the context length. We then weigh each of the coefficients by learned weights $w_{1},w_{2},w_{3} ....w_{L}$, for each of the coefficients of $L$ length DCT coefficients,  and then reconstruct the weighted signal back via IDCT-II to get the signal back, and replace the original signal $e_{i}^{(l)}$. We optimize the architecture according to the loss function, in this case, a Huber loss for a multi-class classification task, which drives the scalars weighing the DCT-II coefficients and the Transformer architecture. The number of tokens was 40, with patch size as 25ms, and the setup of training is the same architecture, with 6 Transformer layers, model dimension as 64 with eight attention heads, as described in \cite{verma2021audio} trained for a total of 300 epochs. We obtain a performance boost to 42.5\% from 41.9\% on accuracy for 1s of audio, with/without using the DCT operation. We add a few thousand parameters to the setup, yielding non-trivial gains on powerful audio Transformer. 
\subsection{Performance for Filtering Approaches And Speedups} 
\begin{table}[ht]
  \caption{Comparison of the negative-log likelihood (NLL) scores (with log to the base e) for our baseline architecture compared with that of incorporating two types of t-f (time-frequency representation) representation: single resolution and multi-scale tf representation with filtering for text-8 test set. (R) We first obtain the speed-up by tweaking the architecture to allow for multi-scale representation filtering. Comparing the performance with doing token-varying filtering.}
    \begin{minipage}[t]{0.45\linewidth}
    \centering
    \begin{tabular}{|c|c|}
      \hline
      Comparison For Performance & Test Loss  \\ \hline
      Baseline Architecture  & 1.04 \\
      Single Resolution T-F Representation  & 1.02 \\
      \textbf{Multi-Scale T-F Representation}  & \textbf{1.01} \\ \hline
    \end{tabular}
    \label{tab:table1}
  \end{minipage}
  \hfill 
  \begin{minipage}[t]{0.45\linewidth} \vspace{-0.88cm}
    \centering
    \begin{tabular}{|c|c|c|}
      \hline
      Configuration & SpeedUp\\ \hline
      \textbf{Baseline v Multi-scale} & \textbf{44\%} \\
      Multi-scale vs Token Varying  & 26\% \\
      \hline
    \end{tabular}
    \label{tab:table2}
  \end{minipage}

\end{table}

We take our baseline architecture as described in section 4 and compare its performance with that of first filtering with $M$ dimension time-frequency representation and to that of learning multi-scale representation. We obtain a significant performance boost by adding a minuscule number of parameters. To give specific context, from \cite{text8-paper}, typically, a boost of 0.01 is significant enough. Adding to that, a jump of 0.04 was obtained going from a 16-layer architecture to that of a 64-layer one as reported in \cite{al2019character}. Learning multi-scale representation followed by filtering performs much better than equivalent $M$ dimension single-scale representation carrying forward the same findings for ASR \cite{zhu2016learning}. In addition, we find that the convergence of our proposed architecture is also faster compared to the baseline model, with speedups 46\% faster with the multi-scale architecture compared to the baseline model. Further, by incorporating the idea of varying filters across token dimensions, as proposed in Section 4.4, we obtain speed-ups in both the convergence in terms of time steps and performance boost over the multi-scale version. 

\subsection{Learned Kernels And Path Interpretation} We open up the first layer convolutional kernel for multi-scale representation and find that we learn kernels that are not sinusoidal, as seen in Figure 3 (C). This re-affirms our choice of learning kernels from scratch instead of using sinusoidal signal-based t-f representation. We plot kernels for each set of convolutional filter lengths of 3, 7, 15, and 31 in Set I-IV. Upon inspection, we find many kernels that are learned to be representative of onsets or sudden late bursts. One of the reasons might be the path-based interpretation, as we have drawn in Figure 1. Each intermediate embedding moves across a token dimension in some high-dimensional space. Our method learns kernels that decompose the path or trajectories in that high dimensional space and filter or constrain them. A high onset-based kernel in the first layer can indicate constraining or allowing the embeddings to move only at a specific rate or to be only constrained in some areas of the embeddings space. We see that using time-varying weights performs better than the fixed-weight LLM. Now, instead of learning to filter globally over the path over the entire token dimension, we can learn adaptive filters locally at every token, with filter characteristics optimized for the current token of interest.

\section{Conclusion and Future Work} In this paper, we have shown how we can draw ideas from traditional signal processing to improve the pre-training of large language models. We find signals in the embedding dimension of every intermediate decoder layer across the token dimension. By drawing parallels to a filter-bank interpretation of the Fourier Transform, we first learn a time-frequency representation similar to a Fourier representation in the space spanned by signals in the intermediate latent representation. We then show how we can do primitive signal processing operations on this learned time-frequency transform, namely filtering. We learn how to weigh this learned transform and reconstruct the signal back again. We learn the time-frequency representation and the weighing function all from scratch, with only the next token prediction driving the entire system. We further improved on our method by mimicking a source separation pipeline that allows us to vary the filter characteristics across tokens on the time-frequency representation learned on intermediate activations, giving further gains. Our architecture operates on intermediate layer representations, making our approach agnostic to the LLM architecture. Finally, we are excited to see more sophisticated algorithms that researchers will bring in a similar theme to the work we have presented to open up the box of ideas in signal processing that will be helpful for generative AI.

\section*{Acknowledgements}
This work was supported in part by the National Science Foundation (NSF) under Grant DMS-2134248; in part by the NSF CAREER Award under Grant CCF-2236829; in part by the U.S. Army Research Office Early Career Award under Grant W911NF-21-1-0242; and in part by the Office of Naval Research under Grant N00014-24-1-2164.

\bibliographystyle{apalike}
\bibliography{ref}
\end{document}